# Modeling belief systems with scale-free networks

Miklós Antal[a] and László Balogh[b]

[a] Budapest University of Technology and Economics, Department of Environmental Economics, H-1111, Sztoczek u. 2, Budapest, Hungary, antalmi@gmail.com

[b] Budapest University of Technology and Economics, Department of Theoretical Physics, H-1111, Budafoki u. 8, Budapest, Hungary, blaci947@yahoo.co.uk

Evolution of belief systems has always been in focus of cognitive research. In this paper we delineate a new model describing belief systems as a network of statements considered true. Testing the model a small number of parameters enabled us to reproduce a variety of well-known mechanisms ranging from opinion changes to development of psychological problems. The self-organizing opinion structure showed a scale-free degree distribution. The novelty of our work lies in applying a convenient set of definitions allowing us to depict opinion network dynamics in a highly favorable way, which resulted in a scale-free belief network. As an additional benefit, we listed several conjectural consequences in a number of areas related to thinking and reasoning.

## INTRODUCTION

Perception and abstract thinking are core areas of cognitive research with extensive literature on fundamental models of human cognition. In the current article we confine the discourse to theories about abstract thinking. The system we developed aims to give account of conscious opinion-arranging processes. For a clear presentation of our model first we have to draft relevant traits of two major knowledge representation theories: the classical propositional model (see e.g. Pylyshyn, 1973) and the connectionist alternative (Rumelhart and McClelland, 1986, for a review see Clark, 1993). Throughout the introduction we will indicate similarities and differences of the "historical" models and our conception.

The well known classical propositional idea considers knowledge as a list of statements. Other types of knowledge like pictures or skills are omitted. This approach has widely been criticized and raised heated debates since the inception of modern cognitive science. We do not interfere in disputes about existing types of knowledge: we ask questions about opinion systems that are characteristically propositional. Our model deals with concrete statements: factual, emotion-based or other types of beliefs that we can represent with sentences are subject to our investigations. Obviously, we analyze a very high level of human cognition (similarly to e.g. artificial intelligence research) by scrutinizing only propositional belief systems and their evolution.

Similarly to classical investigations and controversially to the uniform connectionist system, we start the analysis when environmental inputs are translated to statements. All inputs are considered homogeneous in the sense that there is no distinction between direct knowledge (about tangible objects) and indirect knowledge (about intangible, abstract objects) (Russsel, 2001).

Statements in our model are organized into a network. While statements considered true are the points, links are logical connections or associations that are either positive (+1), negative (-1), or neutral (0). These weights are the only attributes of the undirected links.

Rules of the structuring (automatic processes like in connectionist networks) are given: linking takes place on a probabilistic basis (for the need of a probabilistic system see Pléh, 1998). Points with a great number of connections strongly attract new links. The evolving network structure affects the way new statements are integrated or rejected and the further evolution of the belief network. Linking processes are decomposed into time steps. The stressed importance of network structure and time may recall connectionist theories, while the sequential mechanism used (single processes in time flow) is similar to the classical propositional idea.

We agree that symbolic and connectionist representations complement each other (Eysenck and Keane, 2005). While classically knowledge was conceived as a list of statements and connectionists contended that it was encoded in network patterns (and points were deemed to be meaningless), we claim that it is fruitful to use a network of statements for a representation of opinion systems.

Let us declare at the outset that our model is a theoretical construct. There is at present no unequivocal proof for its relevance that will satisfy all skeptics. Nor is it obvious what "conclusive" evidence could be obtained. Although we accept that none of the examples by itself proves the existence of the phenomenon, we hope that when they are taken together – like weak fibers woven into a rope – the total structure will bear weight.

## THE MODEL

Having seen the basic context, we outline the model in two parts. First we draft the main definitions and static parameters, then dynamic parameters and the mechanism of changes is presented.

### Definitions, static parameters

Definition 1: A network is a complex system of vertices (or points) and links.
Definition 2: A vertex (or point) is a statement considered true.
Definition 3: A link is a logical connection or any kind of association.

The first definition is unambiguous, but two short comments can be helpful regarding opinion networks. First it is obvious that each of us has a different network with different points and link structures. Secondly it worth mentioning that if something is not represented in such a network, then the given person has no opinion concerning this information. The other definitions need some further explanation.

Vertices are simple statements; a compound statement is represented as more simple statements linked together. Vertices may contain any kind of information: facts and beliefs are handled in a uniform way. (Practically, it is not easy to distinguish facts and beliefs, provided we may talk about facts.) We claim that our system involves "local truth" as a driving force (this is in fact soft relativism in cognitive science, for further details see Meiland and Krausz, 1982). That is the reason for using belief systems and opinion networks as synonyms. The conduciveness of this approach can be supported by experimental studies: a great example of reasoning fallacies, the "myside bias", and interactions between opinions and factual information is the article of Macpherson and Stanovich, 2007.

Links may be logical connections: one statement is a consequence of another, two statements are contradicting etc. Another possibility is that links are built on an associative basis: two statements have similar topics, subjects, subjectives etc., or there are emotional liaisons, grammatical similarities, even sub-symbolic connections.

In a static case, vertices are characterized by their degree parameter:

1. Degree: $k_i$ - the number of connections of vertex $i$

Links are characterized by their one attribute:

2. Weights +1, 0 and -1 show whether the linked vertices are in positive, neutral or negative connection. (This scale can be made more precise in a later version of the model.)

A short description of these factors may be useful here:

Degree is simply the number of statements connected with the given vertex. A central statement is connected to a huge number of other statements; peripheral statements are linked to only a few others.

Links are positive, negative or neutral: two vertices are more solid together (+1), they rather impair each other (–1), or they are independent (0) like two "facts" about the same topic. Positive links strengthen the network; points help each other to remain in the system. Negative links stress the network and act towards a collision. These effects will play important role in the system's dynamics.

## Dynamic mechanism and parameters

As we strive to give account of dynamic processes and use simulation results of a computer code (available at http://home.fazekas.hu/~blaci/belief_networks/), a correct presentation of dynamics is inevitable. The main definitions of network dynamics are the followings:

Definition 4: An input is a new point for the network (with non-existing content).
Definition 5: At a certain time one and only one point of the network is active (it has a distinguished role in dynamic processes).
Definition 6: A time step is a discrete time interval for elementary changes in the network. (Detailed elucidation is given below.)
Definition 7: In every time step $n$ links randomly vanish. (This random process can be interpreted as forgetting (Bednorz and Schuster, 2006).
Definition 8: A vertex losing all its links vanishes.

Main dynamic parameters driving all processes:

3. Compatibility factor of a vertex: $g_i$ - gives the probability that the given vertex is in positive (strengthening) connection with a randomly chosen vertex - a number between 0 and 1
4. Contradiction factor of a vertex: $h_i$ - gives the probability that the given vertex is in negative (weakening) connection with a randomly chosen vertex - a number between 0 and 1

5. Fitness factor of a vertex: $f_i$ - shows how much a vertex takes part in linking processes (compared to other vertices with the same number of connections) - a number between 0 and 1. (If $f_i = 0$ then this vertex never makes connections, if $f_i = 1$ then it is maximally capable of linking.)
6. Negativity tolerance (consistency) of the network: $H$ - shows what proportion of the connections of a certain vertex can be negative - a number between 0 and 1, global parameter. If the proportion of negative links is proved to exceed $H$, the vertex is ejected.

Some remarks about these factors:

Compatibility and contradiction factors show how much a certain point fits in the network: if we believe in something and our network treats an inconsistent point, than $g$ is small and $h$ is big. There are neutral connections, so $g_i + h_i = 1$ does not hold for every $i$.

Fitness factors allow "newcomers" to become richer in links than elder points. If one point has $k_1 = 5$ links and fitness factor of $f_1 = 0.1$ and another vertex has $k_2 = 1$ link and $f_2 = 0.5$ then an input is linked to each of them with an equal probability. If we did not use fitness factors, then the older vertices would always dominate the networks. The importance of older vertices holds true even by the usage of fitness factors, but in this case changes in the order of significance are permitted. For a correct mathematical description see Appendix A.

Negativity tolerance is a crucial factor: if $H = 0$, then no contradictions may occur, just as in the network of some mentally ill people. On the other hand, $H = 1$ resembles the case of schizophrenic belief systems.

Here we point out that there are two different ways of vertex ejection in the model: one due to the loss of links and another due to an inadmissibly high ratio of negative links.

Having defined all the needed notions and parameters we are ready to delineate rules in opinion networks. These rules impose different kinds of changes: new links are formulated other links vanish, points are integrated others losing all their connections disappear. Occurring processes are deemed to depict the way we organize our opinion structures.

Development always takes place in the vicinity of an active point: linking and checking procedures start there causing vertex integration and/or ejection. An active point is considered to be a statement one is currently thinking about.

There are two mainly different cases: input processing, when a new point containing unknown information is built in; and active point processing that is the general case for network structuring starting from already existing active points.

Input processing starts when an input arrives and takes activity. (Here we see that the notion of active point includes the one of input: all inputs are active points for a certain time.) The first step of input processing is preferential attachment: links are established between the input and vertices of the network. The probability of the formulation of a new link is directly proportional to the degree of the existing vertex and to its fitness factor. If all links are built of an input (an input carries a given number of links), then the types of the established links are decided in a second step in accordance with the input's compatibility and contradiction factors. A consistency test is run in a third step. It is checked whether the ratio of negative links does not exceed the negativity tolerance limit ($H$) for any of the points. If there is a vertex with an unadmittable proportion of negative links, then it is ejected. Special cases and a possible chain of tests are elucidated in Appendix A. As the sum of link-weights controls changes regarding statements considered true and this sum is decisive whether statements remain in the network or they are ejected we may speak about "local truth" as a driving force.

If a point (the former input) is linked in, it becomes a point of the network. If the point is still active (that is time-dependent), there is a further linking process. The mechanism to treat existing points in the network (viz. thinking processes) is the following. A two step random walk on the network starts from the active point. Random walks are weighted with the fitness factors, i.e. the probability to reach a certain neighbor is proportional to its fitness factor. We reach a vertex and link it with the active point. (One time step is needed till this point.) Then there is a decision based on the compatibility and contradiction factors of the input, whether the link is positive, negative or neutral. Then comes the consistency test. (Ending in one time step if there are not too many negative links and no ejection is needed but consuming much time if a chain of tests is needed due to vertex ejections.) Two step random walks, linking and consistency tests are repeated till time runs out (e.g. a subsequent input arrives). According to the scale-free structure and small world property various formulations may grow up, and time devoted to a vertex highly influences its future role in the network. Details are elucidated in the next section.

## DISCUSSION AND APPLICATIONS

### Network features

In this section we analyze networks given by the former mechanism from a structural point of view.

First, we ascertain that there is very a special parameter setting: if all fitness factors are equal, all links are positive, no time is given for random walks (time is devoted to consecutive input processing procedures), and there is no random edge removal (forgetting), then we obtain scale-free distributions (Barabási and Albert, 1999). The degree distribution of a scale-free network is a power-law decay (details in Appendix B). If we use logarithmic scales, we get linear decay. (Throughout the article logarithmic plots are used for degree distributions.) As scale-free distributions are central to our investigations we show that if we do not use any of our parameters, we get back the original Barabási model (Fig. 1a), and that scale-free structure is kept even if we use all parameters (Fig. 1b). In the latter case inputs had more links, uniformly distributed fitness factors, and equal chances for contradiction, compatibility and neutrality ($g_i = h_i = 1 - g_i - h_i = 1/3$), moreover time was given for linking processes. Negativity tolerance was chosen to be $H = 1/2$ and even random link removal was present. (Parameter settings and details about all figures are given in Appendix B.)

**Fig.1.** Scale-free degree distributions

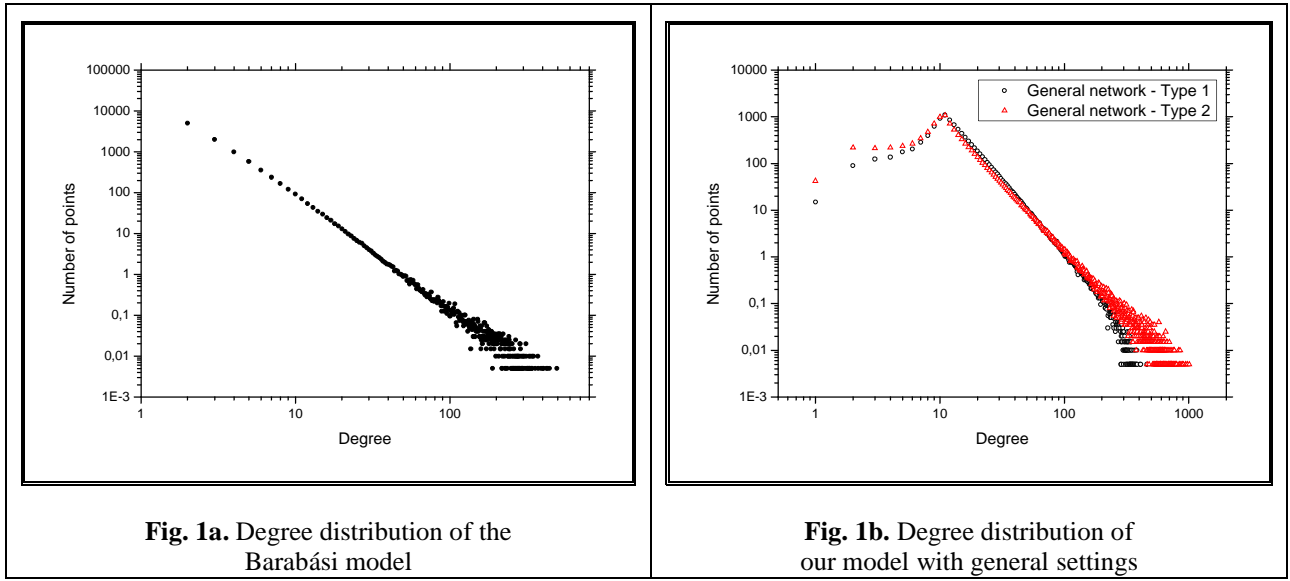

**Fig. 1a.** Degree distribution of the Barabási model

**Fig. 1b.** Degree distribution of our model with general settings

These simulation results prove that the two structures are essentially the same. Consequences of this observation are far-reaching since preferentially built scale-free structures have special characteristics as outlined in the followings.

*Small world*

The first structural feature resembling common experience about belief networks is small worldness. It is an everyday observation that associations in our mind may lead very far in a few steps. In terms of networks this feature is called "small world" property. The diameter (average shortest distance between two randomly chosen points) of a small world network is incomparably smaller than the number of points, the order of magnitudes widely differs (Albert, Jeong, and Barabási, 1999). The small world characteristic makes an extremely diverse flow of thoughts possible. Thus, we expect a model encompassing small world attribution.

**Fig. 2.** Diameter as a function of network size

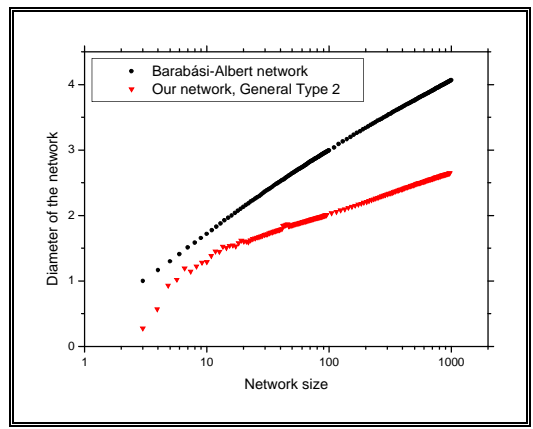

Fig. 2 shows simulation results: the average distance between two points in a network of 1000 points is approximately 4,1 in the original Barabási network and 2,7 in a general version of our networks. Clearly, in keeping with the expectations, the model produces small world networks.

*Scale-free network*

The second expected feature given by simulations is scale-freeness. The distribution itself means that the number of statements of a given importance obeys a power-law. (Here we note that importance and the degree of a point are not equivalent e.g. because it is also interesting how central they are concerning walks on the network, though, to a first approximation we use degree distributions to capture importance.) No single supreme thought is present in a healthy mind and the few very important core statements are closely followed by others. We can always find more and more statements of slightly smaller importance till we arrive to the most populous periphery. Scale-free distribution implies that opinion systems obey Pareto's 80/20 law. As expected, the majority of time is devoted to a minority of statements in our networks.

Scale-free structures are robust: if a randomly chosen point is removed, it usually does not affect system behavior, as disappearing points are usually peripheral. However, "error tolerance comes at a high price in that these networks are extremely vulnerable to attacks (that is, to the selection and removal of a few nodes playing a vital role in maintaining the network's connectivity)" (Albert, Jeong, and Barabási, 2000, p. 378). We argue that our belief systems work in this way: the loss of peripheral statements does not mean much for the network, but attacks against core opinions may ruin the system causing serious psychological problems. (If conceiving thinking as a random walk on a network of thoughts, we always encounter routes crossing large centers; if they are attacked, a number of walks are spoilt.) As our networks are scale-free, we obtain error tolerance and attack vulnerability.

The fact that we imagine opinion systems as preferentially evolving scale-free networks (Barabási and Albert, 1999; Barabási, 2002) should not be stunning for several reasons. First, it is shown that words in human language linked by co-occurrence in sentences form a scale-free network with small world characteristic (Cancho and Solé, 2001). Secondly, small world and scale-free properties also appear in conceptual networks where similarities of concepts connect words of a language (Motter, de Moura, Lai, and Daspguta, 2002). Moreover, the same features hold for cognitive maps (Özesmi and Tan, 2006). If we conceive texts as linearized versions of subnetworks, we may also refer to strong correlations between text quality and complex network features (Antiqueiraa, Nunesa, Oliveira, and da Costa, 2007). Finally, the principle of preferential attachment also seems to be reasonable: people associate to statements that are strongly represented in their networks.

*Anomalies from scale-freeness*

Though, scale-free structures do not prevail in some cases: in the beginning when the network consists of a small number of points, our structures rather resemble random networks. On the contrary, if a point with a high fitness factor is present star shaped networks may occur. Such environment dependent transitions are generally observed in networks (Derényi, Farkas, Palla, and Vicsek, 2004). The former case (random network) can be interpreted as an immature, not well structured system that is characteristic for the inception of development processes. (Apparently, a small number of points can not form a scale-free degree distribution due to statistical reasons, but as the number of points grows scale-free distribution emerges.) The latter (star shaped network) is something completely different:

there is a statement of unique importance in a network. This leads to a conformation that determines behavior: the exceptional point gathers a large number of links, most random walks go that way, and that point will be the absolute center as shown in Fig. 3. (The peak in the right is not a single point with a probability of 1 but approximately 100 points close to each other with probabilities of approximately 0.01, as the average of 10000 simulations is depicted in the figure. Colors indicate different simulations: the ordinal number of the special point was modified from 1 to 32.)

**Fig. 3.** Degree distribution of a star shaped network

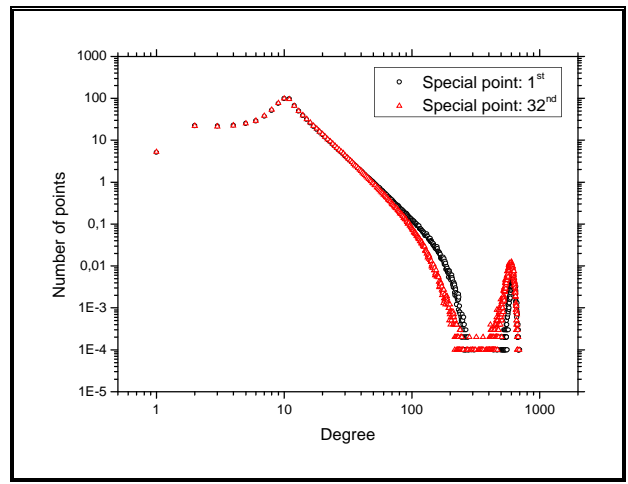

What could it mean in reality? As random walks cross the exceptional point extremely often, a star shaped structure hampers sufficient thinking. Instead of reaching vertices more or less proportionately (e.g. according to a scale-free distribution), we always get back to the center. Vertices of lower degrees are unlikely to be linked, normal system behavior and structuring are inhibited, and significant changes are improbable. This "polarization" can be observed in many areas as pointed out by Lord, Ross and Lepper in 1979. Politics, racism, and private life are all fields of star network conformation. Often, there is absolutely no chance to integrate certain statements in a network, see political views. Too strong (usually emotional) centers lead to a grotesque case: for instance people evaluate information in the mirror of political parties and not the parties in the mirror of information. (It is shown that emotions play a decisive role in political reasoning, see Westen, Kilts, Blagov, Harenski, and Hamann, 2006) This is a typical devastating effect of a star shaped subnetwork: new information are connected to the center and only allowed to remain in the network if there is a non-negative link between them. Similarly, there are conflict zones in private life: we know which part of the network should not be activated so as to avoid conflicts. Usually, star shaped structures are problematic parts of opinion networks.

## Inherently encompassed phenomena

A major advantage of the outlined model is that it inherently encompasses phenomena emerging in a diverse range of everyday life. In the followings we show particular behavioral characteristics of the model that can be matched to observations of reality.

*Crucial early points*

First, it is a common experience that statements accepted in an early phase of opinion system formulation are of huge importance. In other words, first stimuli have a massive effect on our future way of thinking and it is not easy to remove old, entrenched ideas from belief systems. Here we may refer to upbringing of children and the stressed importance of early inputs largely determining mentality (Dawson, Ashman, and Carver, 2000). It is often argued that lots of psychological problems stem from early ages – when incorrect centers are built in, we claim. As a smaller scale example: if we first meet someone and thus a new part of the system arises, first impressions have great importance. In our model all these effects are deemed to be manifestations of network-evolution based on preferential attachment, where early vertices are of great importance, being located at the high degree end of degree distribution (Barabási, 2002). For an illustration, see Fig. 4.

**Fig. 4.** Degree of points by their sequential order

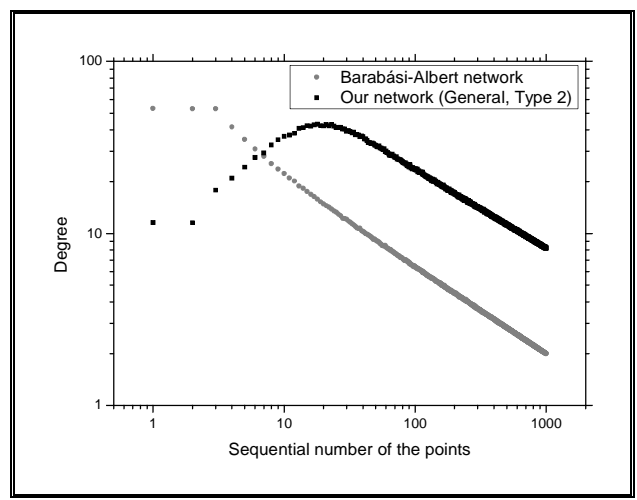

*Time as a determinant of importance*

Second, it is unquestionable that time plays an essential role in the formulation of belief systems. Advertisers try to capitalize the fact that the more time is given to process an input, the bigger is the probability that it gets integrated and becomes a center. Besides, people who are rarely stimulated (and thus have much time for each input) are greatly affected by the few stimuli, these vertices become centers. These effects are included in our model: the more time is given to a vertex, the more connections it will build and the higher degree it will reach. Extremely long processing times lead to extreme degrees as shown in Fig. 5. Apparently, the peak refers to the high degree of a single point while other points have much lower degrees. (Again, the peak in the right is not a single point with a probability of 1 but approximately 100 points close to each other with probabilities of approximately 0.01, as the average of 10000 simulations is depicted in the figure.)

**Fig. 5.** Degree distribution of a star shaped network
with extremely long time

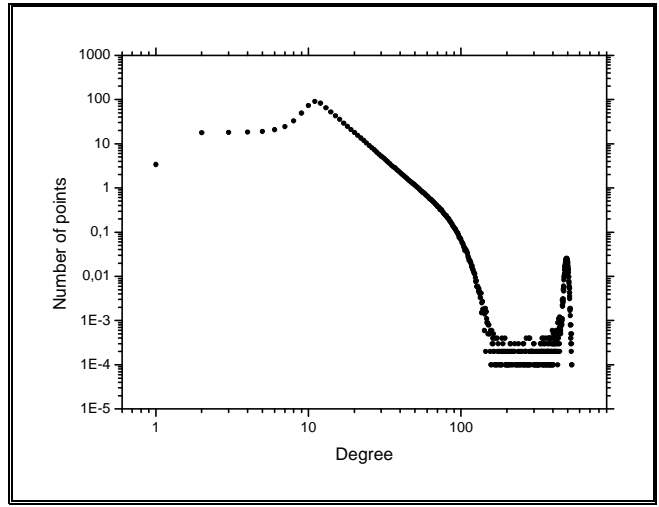

*Size as a measure of robustness*

In a third section we analyze possibilities of changes in opinion structures. There is an enormous difference between statement integration chances if developed and undeveloped belief networks are juxtaposed with one another. New ideas may swiftly achieve great significance in an immature network but are not likely to lead to drastic changes in massively diversified, highly developed structures. Children and illiterate people are strongly exposed to fanatic ideas while academic professors usually do not commit suicide attacks. Children are gullible while old people are sometimes unable to integrate new information. These are natural consequences of network size in the model. Once again, drawing parallel between significance of a statement and its actual place in the degree distribution (how many links does a point have compared with the others) we can assert that points (e.g. with a relatively high fitness factor) reach higher levels of significance more easily in networks containing less points and edges. This effect is represented in Fig. 6. By smaller sizes, the "attacker point" can achieve maximal degree in the network while by greater sizes the maximal degree is significantly larger than the attacker's degree. (Please note that we use logarithmic scales.)

**Fig. 6.** Influence on different sized networks

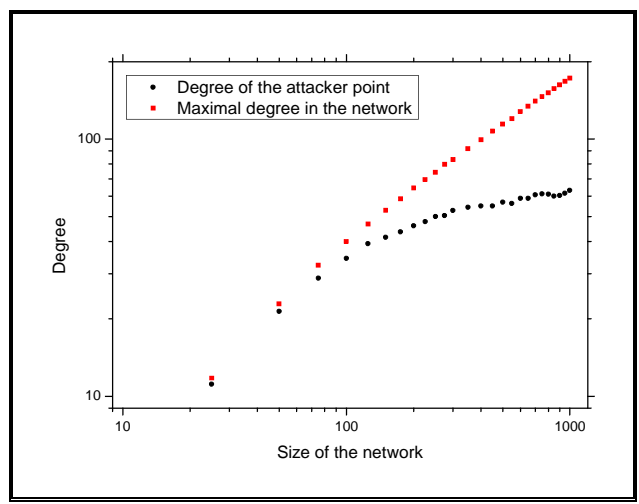

Elder, highly qualified people usually have more developed networks as it follows from the previous arguments about the role of time, so their degree distribution is wider, they have more vertices with large numbers of links. Obviously, it is not easy for newcomers to attain such high degrees what is an explanation for the above mentioned experiences. On the other hand, networks with a smaller number of vertices and less connections are more easily affected by novelties. Though, there are a number of different ways of change that are under study in the following three subsections.

*Learning – optimal input frequency*

Learning for instance is a changing mechanism of pivotal importance. While classically it was deemed to be the sheer enlisting of a new statement and connectionists described learning with changing weights of links, we combine the two approaches. The appearance of new statements and the construction of links (viz. structuring) jointly explain the way we learn. Our model precisely reproduces some nontrivial observations.

Again, we start from a large scale example. It is well known from international surveys that Prussian school systems, where a comprehensive knowledge is offered and large amounts of facts are taught (so there are lots of inputs) produce an excellent elite class and a poor average (OECD, 2004a). That can be underpinned by the model behavior: the complexity of an evolving network heavily depends on the linking capability of the student. (This can be interpreted as the real time equivalent of a time step in the model: those who learn or think faster need less time in reality to perform steps of linking and checking procedures.) Without sufficient linking capabilities information is useless, they form rapidly vanishing islands. Further information have no vertices where they could link to, the network does not improve. That happens to most children in a Prussian-type school: they just do not have enough time for structuring. (The previous quite general statement pertaining to overall performance relies on the fact that e.g. text understanding – that is clearly strongly related to linking capability – is remarkably correlated with overall performance (OECD, 2004b).) In contrast, sufficient linking capability plus a huge amount of vertices expedites structuring: the number of possible links rises very fast with a growing number of vertices allowing optimal development. Reflecting this case differentiated education is introduced in several schools: learning (linking) methods are taught for those who require it and information for the others who are ready to integrate.

Here we reach a smaller scale problem: similarly to school systems, efficiency of individual lectures is by large determined by its speed. Frequency of inputs (the amount of information given in a time period) determines performance. Our model gives account of this feature: starting from a given network, working with nonzero random link removal and fixing the number of time steps available there is an optimal number of points to be given in the time period to reach a maximum number of integrated vertices after the process. The number of points in the network after the learning process is depicted in Fig. 7. The original network consisted of 1000 points and 1000 links (2 links for each point on average); the number of added points varied between 10 and 100, the number of available time steps was fixed to be 1000.

**Fig. 7.** The effect of input frequency on learning

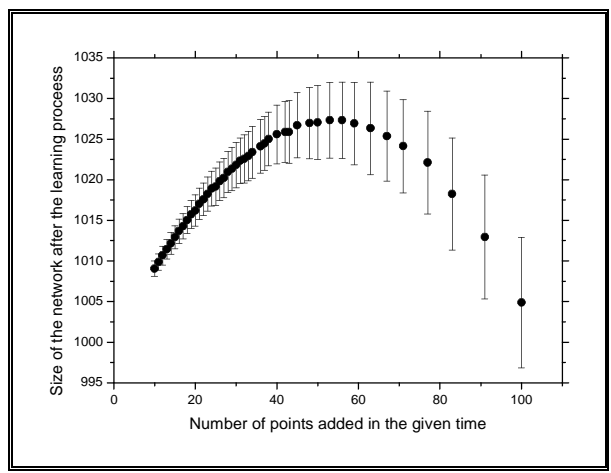

Obviously, it is worth building more connections if there is a danger of losing access paths due to forgetting. Moreover, the constructed topology determines resistance against random link removal. If we build a linear network with statements linked only to the subsequent statement as it often happens in history lessons, then large parts of the curriculum may be unreachable in the network due to the loss of certain connections. It is an everyday observation that we forget everything about some former studies and once being reminded of a certain statement we are able to bring up a few connected statements but then we are stuck again; the system is not integrated as a whole. Arguably, the curriculum structure is very important. Perhaps robustness of scale-free networks could be exploited so as not to lose access paths so fast. (As a matter of course, the problem and the need for appropriate structures are recognized without such theoretical foundations.)

In addition, we may refer to exams and particularly oral exams as examples of the usefulness of network based thinking. Teachers usually try to roam through the network of the student so to check the existence of certain points and connections. This is a reason for stressing the importance of links and the structure as a whole in contrary to the barren subsistence of vertices. We contend that understanding is hidden in the integration process.

*Restructuring in debates*

Another often encountered type of change where people try to shape the other's network is debating. In a dispute the goal is to build a strong system (a network) of own arguments and to destroy the network of our opponent. The latter is done by causing percolation of the opponent's structure by building in as many negative links as possible. There are different means how we can achieve this: we may point out contradictions of the structure, integrate new vertices for establishing negative links between existing vertices or integrate new vertices that are in contradiction with existing vertices themselves. It can be useful to draw a network of the opponent's arguments so to analyze it and find the ideal vertices to attack or vertices that are not worth considering (e.g. peripheries that percolate after an attack). The frequently applied technique to simply confute all the statements with one argument is far from optimal. The same means can be used when defending our own network under attack. This method can also be a valuable tool of evaluating debates.

A further application of such a representation of debates arises from the fact that the center of a debate – the topic – is often unequivocal. If we drew up lots of networks of high quality argumentations then we could evaluate existing indicators of the centrality of a vertex in a network.

*Subnetwork integration – manipulation or discovery*

Finally, some interesting experiences about manipulation and scientific productivity. The model allows a very special way of vertex integration: if a new part of the network evolves separately from the former parts of the network and only a few connections are built between the two parts, then it is possible that contradictions remain undiscovered until enough time is given for thinking about the new points. This is certainly the case of urban legends and conspiracy theories: a vast amount of new information is delivered with a few obvious connections to reality and the theory itself is a positively linked network. A nearly disjoint structure of points strengthening each other does not allow vertices to be dropped. This strategy can also be used in a persuasion to get our information across without being rejected (e.g. due to a star shaped subnetwork) and this may lead to changes in the original network if we manage to build such viable structures that can override formerly developed parts.

If two distinct substructures are not controversial but connections are unnoticed, we may talk about "local discoveries" when connections are finally built. If local discoveries uncover unknown relations between two research areas, then we may produce scientific results. The magnitude of restructuring thus follows scale-free distribution due to the distribution of the size of connected parts. This means that most useful scientific ideas are distributed unevenly: a researcher produces the majority of his results in a minority of the time devoted to the job.

## Further implications and conjectures

In this section we demonstrate applicability of network theoretical notions for belief systems then point out to the potential of the model to interpret widely used but vaguely defined everyday notions. This chapter does not include systematic simulations, so arguments are rather conjectures for future studies.

*Stability in structure and functioning*

Noise is an external effect causing changes, possibly destruction in a network; noise filtration is a mechanism to avoid dramatic harmful changes. Self-organizing evolutionary networks always have methods to resist such changes (Csermely, 2006). In our case noise is coded in inputs with high contradiction factors, its filtration is tackled by the negativity tolerance factor, the modularized structure itself (destructions can be localized) and perhaps by protecting modules (consciously giving negative links to certain inputs).

In general, diversity of behavior emerges if the number of links decreases. In our model it means that a great number of links enable associations to reach local centers in a few steps as the small world feature takes shape. In the lack of a sufficient number of links behavior becomes highly dependent on the structure defined by the existing links, behavior will not be averaged by the densely linked conformation. Indeed, unexpected reactions are characteristic for people who have undeveloped networks.

However, it is observed that too densely linked structures are also vulnerable (Watts, 2002). This phenomenon is also encompassed in the model: if a vertex drops out and another is ejected due to the loss of the first (to which it was positively linked) then there will be a high probability that some vertices loose two positive partners and have to be dropped. If the network is too densely linked, the process can result in system-level destruction. (Such a process can be generated with the computer code, although, there is no simple way to include results in the article.)

*Psychological and communication problems*

If a network is exposed to abounding new information containing inputs with relatively high contradiction factors then checking procedures may be interrupted by new inputs leaving inadequate points in the network. This lack of enforcement of rules in the network can lead to a feeling firmly associated with cognitive dissonance. More generally, psychological problems are often related to the fact that our own rules are not vindicated. If there are forbidden parts of the network containing unacceptable proportions of contradictions, then these locked up problems can cause psychological malfunctions. Psychologists often do not really intervene in the development of belief systems but they lead the patient to certain problematic areas of their own network.

Also, there is a possibility to interpret communication problems like failed talks. If partners do not want to follow the routes dictated by the other's speech and only perceive single inputs or activations from it, then there will be no real conversation: both speakers roam their own networks.

*Creativity and humor – distant linking*

Intelligence and creativity are notions definitely included in the scope of the model. If we think about intelligence as a quantity measured by IQ tests, then it is a kind of problem solving capability where two main features are required: having well-shaped local, small-scale statement structures on the one hand and being fast in searching on the other. In contrast, creativity is an ability of distant linking or more precisely, we call someone creative if his network is well-structured on a larger scale with sufficient connections between otherwise disjoint subnetworks. These definitions explain why intelligence and creativity are correlated till a certain IQ value (about 115) and become independent above. Fast search in confined areas help problem solving on a larger scale as well. It is needed in creative problem solving to reach vertices that are a few steps away, i.e. before or after using the "creative link" between the distant areas. Although, no matter how fast we are in local search there is no real chance to find connections between two distant points without sufficient creative links because after a few steps there is an astronomic number of possible routes that can not be checked by a "brute force" technique. Given an eligible speed of search (depending on local structures and rapidity) the determining factor in creative problem solving will be the existence of far-reaching creative links.

The observed connection between humor and creativity is also originated from this point: distant linking appears in humor in most cases – the punch line is usually a statement from a completely unexpected part of the network. A sense of humor thus relies mainly on two factors: the advanced state of the used structures (not all kinds of jokes are equivalently understood by people) and the ability of distant linking.

## Conclusions and perspectives

In the present article we delineated a model of belief systems with a potential that can be harnessed in a wide range of research areas. The sheer structure of opinion networks, changes determining evolution, and specific behaviors that are given by the model have relevant implications regarding a number of cognitive psychological processes. Naturally, we are far from a proper description of opinion system formation and development, but the usage of scale-free network theory for modeling statement networks is promising.

There are some obvious extensions of the method making specific properties or descriptive features more precise but complicating the model on the other hand. First, weighting of connections can be refined to give a nuanced picture of binding strength between statements. Though, a weighting mechanism is to be defined then. A possible solution can be to relate weights with usage frequency (like in several connectionist PDP models). Secondly, activation spreading can be included in the model. In the lack of inputs activation may spread on the network enabling more complex structuring processes. Thirdly, points may be characterized with an additional factor – call it color – that refers to its topic including features that are relevant in linking (object, emotions, grammatical form etc.). All relevant features give one color to the point. If an input comes (with given colors), then linking starts with a probabilistic decision about the color that will be used when building the connection. The following step is the one we used in the original model applied to vertices that are marked with the given color. Consequently, points with more common colors (viz. stronger similarity) are linked with a bigger probability. Such modifications may improve the effectiveness of the model in several areas.

Apparently, there are scores of other possible improvements out of which we mention only one here. Networks are sensitive to drastic changes. A factor showing the magnitude of changes in a given time period tells a lot about the mental state of the person. It could be analyzed how certain environmental circumstances (frequency and type of inputs) affect mental status. The role of the original network may also be of crucial importance.

There are some questions that will determine the future of this model: exact methods for network mapping and quantifiable tentative steps for further substantiation are surely such. Still, without answering these questions some applications are ready to be tried and perhaps the approach towards opinion structure research is expanded in a way.

We hope that a proper guidance was given to roam through a statement network about belief systems and researchers are inspired with properly fitting inputs. If new connections arise in the integration processes developing the structure of knowledge about belief systems, then this article attained its purpose.

## ACKNOWLEDGEMENT

We thank Prof. Peter Csermely for valuable comments and careful proofreading.

# Appendix A

## Mechanism of the model

**Fig. 8.** Flowchart of the algorithm

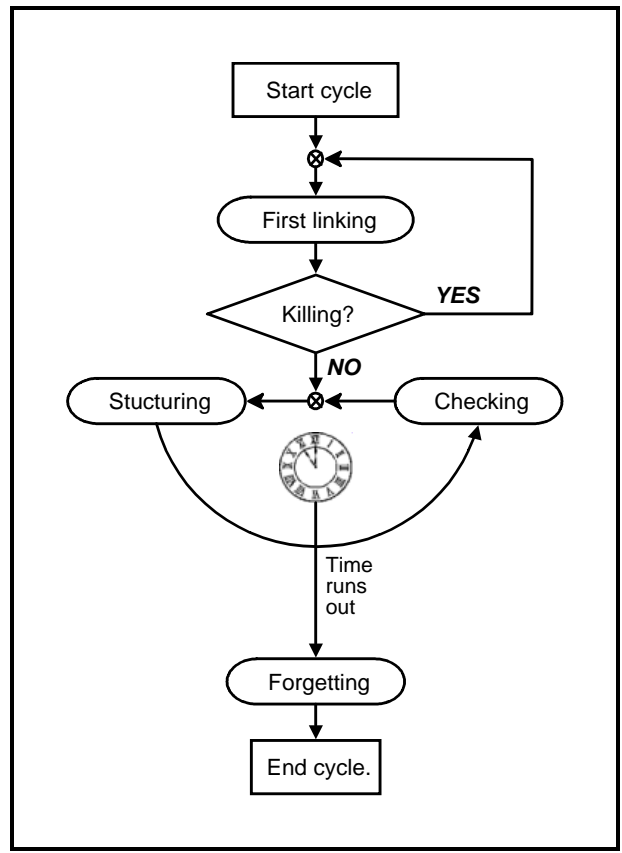

Here we explain the mechanism of our model in details (see Fig. 8). The computer program from which simulation results are obtained uses exactly these definitions and algorithms. A short summary is given regarding structural consequences.

We realize network construction in a series of cycles. In each cycle the system processes only one input point: establishment of new connections between the point and the existing network is endeavored. According to the parameters it will succeed or not. If the input point joins the network it induces further linking until a new input arrives. The main units of the process are shown in Fig. 8.

There are three parameters in the cycle process denoted by $U$, $E$ and $F$. They stand for the followings: $U$ – the number of edges carried by the input point, $E$ – available time steps for the whole cycle ("time for thinking" about the input information), $F$ – determining the amount of edges to be forgotten (disappearing randomly) in one cycle.

## Adding new points

We create input points (denoted by index $i$ in the followings) with parameter values $f_i$, $g_i$ and $h_i$. (Parameters $g_i$ and $h_i$ may be considered in the following way: the probability for a particular edge to be positive, negative or neutral is $g_i = \frac{a_i}{a_i+b_i+c_i}$, $h_i = \frac{b_i}{a_i+b_i+c_i}$ and $1 - g_i - h_i$, where $a_i$, $b_i$ and $c_i$ denote the number of possible positive, negative and neutral links, respectively. We used these expressions to evaluate parameters $g_i$ and $h_i$ from factors $a_i$, $b_i$ and $c_i$.)

## First linking

As mentioned before new points should follow preferential linking in order to get scale-free network structure. Mathematically it means that the probability of a new edge attaching to a particular vertex (denote this non-neighboring target vertex by $t$) is proportional to $k_t$. Taking into account our extra parameter referring to the attractiveness of points, one can formulate the expression

$$P(\text{"linking to vertex } t\text{"}) = \frac{f_t k_t}{\sum_p{}' (f_p k_p)} \qquad (A.1)$$

$\sum_p{}'$ means that index $p$ runs over all points which are not connected to point $i$, factors $f_p$ and $k_p$ denote the compatibility factor and the degree of point $p$, respectively. The probability of building a positive, negative or neutral link is $g_i$, $h_i$ and $1 - g_i - h_i$, respectively. (If there are no edges in the network i.e. in the very beginning of a simulation one cannot evaluate expression A.1, so the following formula can be used instead: $\frac{f_t}{\sum_p f_p}$, where notations are similar to those used before, but now the sum $\sum_p$ runs over all points except $i$.)

This linking step must be repeated $U$ times. Then we should check whether the new point is consistent enough with the "old" network. This is performed by calling a $Killing(i)$ function (discussed below). If the output of $Killing(i)$ is "YES" – meaning that the new point does not fit in the network – all of its edges will be cleared and the "First linking" process will be restarted. If the output of $Killing(i)$ is "NO" – meaning that there are not too many negative links – operation *Structuring* follows.

In this process time is needed for checking as it is elucidated in the next section (killing). If the available time runs out without attaching the new point, we go on to the next input point.

## Killing

For an arbitrary point $j$, $Killing(j)$ returns "YES" if point $j$ is more inconsistent with the network than the limit value fixed by parameter $H$. $Killing(j)$ returns "NO" if the ratio of negative links of point $j$ does not exceed $H$. Mathematically:

$$Killing(j) = \begin{cases} \text{YES} & \text{if } \dfrac{\#\{\text{negative links of j}\}}{\#\{\text{links of j}\}} > H \\ \text{NO} & \text{if } \dfrac{\#\{\text{negative links of j}\}}{\#\{\text{links of j}\}} \leq H \end{cases} \quad (A.2)$$

where $\#\{...\}$ denotes the number of elements of the $\{...\}$ set, $H$ is the consistency or negativity tolerance of the network.

One $Killing(i)$ test consumes 1 time step.

## Structuring

To construct new edges between the input point and former points of the network, two-step random walks start from the input point. The first step from $i$ leads to its neighbor $n_1$ with the following probability: $P(n_1) = \dfrac{f_{n_1}}{\sum_p f_p}$. (Where $\sum_p$ means a summation over all first neighbors of $i$.) In the next step we arrive to a second neighbor $n_2$ with the probability given here: $P(n_2) = \dfrac{f_{n_2}}{\sum'_p f_p}$. (Here $\sum'_p$ means a summation over all neighbors of $n_1$, except for $i$ itself.)

Then we establish a link between the input and the afore mentioned point $n_2$. The new link will be positive, negative or neutral, respective probabilities are $g_i$, $h_i$ and $1 - g_i - h_i$.

## Checking

After structuring processes it is possible that a point due to a growing number of negative connections does not fit in the network any more. To avoid discrepancy in the network checking mechanisms are needed. First, two tests are called: $Killing(i)$ and $Killing(n_2)$. According to the results of these tests:

1. If none of these two points should be removed: *Structuring* goes on.

2. If input point $i$ should be removed and $n_2$ not: we clear all the edges of the input and restart the *First linking* section. (This can be considered as a new chance for the input to get integrated.)

3. If point $n_2$ should be removed and point $i$ not: we remove $n_2$ and start a checking mechanism to investigate, whether the removal of $n_2$ affected other points as well. (The falling number of positive links may lead to ejection of new points.) Details are elucidated in the next section (Self-Consistency Test).

4. If both input point $i$ and point $n_2$ should be removed: we remove the one with a smaller number of edges and go on with processes described in either case 2 or case 3.

## Self-Consistency Test

This is a test aiming to remove negatively linked points (where *Killing* would result in YES). The test requires a starting point (to be tested first) and time for the process.

When we remove a point, it can happen that a positively linked neighboring point – by losing this positive connection – gets under the required level of consistency ($H$). Therefore we should check each point which is positively linked to an ejected point. We introduce a list (called "blacklist" hereafter) to store the points that are waiting for such a test. A brief delineation of the process is shown in Fig. 9.

**Fig. 9.** Algorithm of the Self-Consistency Test

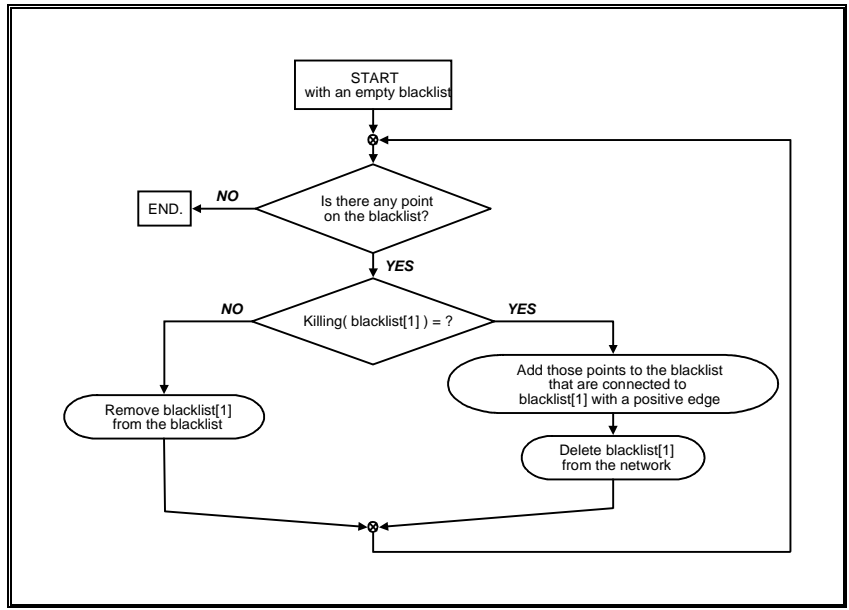

Each Self-Consistency Test starts with a blacklist containing only the first point that induced the process. (A previous blacklist – if there was such – is lost while starting a new test.) We always analyze the first element of the blacklist. (Denoted by *blacklist[1]* on Fig. 9.) We determine whether blacklist[1] is to be removed or not – of course – by calling a $Killing(\text{blacklist}[1])$ function. If the output of $Killing(\text{blacklist}[1])$ is YES, we delete it and put all positively connected points to the end of the blacklist ranked by their *f* values. (One point is put on the list only once – here we refer to the case when it is already on the list when another neighbor is ejected.) If the output of $Killing(\text{blacklist}[1])$ is NO, we remove the point from the blacklist without any further operation and continue with the current first element of the blacklist. (Removing one point from the blacklist does not mean that it would be out of the blacklist forever. It can be put back if other ejections induce this.) This algorithm runs till we get an empty blacklist or till available time runs out. Again, the running of one $Killing(\text{blacklist}[1])$ function consumes one time step.

## Consequences

Evolution rules determine the structure of the evolving network. Preferential attachment leads to a so called scale-free network, viz. where degree distribution obeys a power law: $P(k) \propto k^{-\gamma}$, where $\gamma$ is a fixed number. In case of preferential attachment (where $\gamma$ is usually between 2 and 3) the 80/20 law and the emergence of the small world characteristic are straightforward consequences. Our linking processes involve preferential attachment and further linking based on random walks. This mechanism also produces scale-free networks, as it is shown in the original text of the article.

# Appendix B

## Network features and simulations

In the course of network research degree distribution plays an inevitable role. A degree-distribution diagram shows the number of points with a given number of links. For scale-free structures the probability of having $k$ links is:

$$P(k) = \frac{k^{-\gamma}}{Z_\gamma}, \tag{B.1}$$

where $\gamma$ is the exponent of the distribution, $Z_\gamma = \sum_{k=1}^{N} k^{-\gamma}$ is a normalization factor, $N$ is the total number of points. A conspicuous presentation of power law distributions is possible, if log-log scales are used, since:

$$\log(P(k)) = -\log(Z_\gamma) - \gamma \log(k) \tag{B.2}$$

is a linear function and $\gamma$ is the slope of the line. As the network is built on a probabilistic basis, all concrete networks differ. Usually a great number of networks are built with the same parameter set and degree distributions are averaged to get smoother, more precise curves.

## Simulations

*Figures 1a and 1b*:
The following simulation was performed to prove that our model produces scale-free degree distribution under quite general circumstances. The Barabási model was built with appropriate parameters and a general setting (with no distinguished parameters that could cause special effects) of our model was run. The given parameters are given in Table 1, results are depicted in Fig. 1a and 1b. (RND means a random number between 0 and 1 from a uniform probability distribution.)

To recall the meaning of the parameter we give short explanations for the letters:
$H$ : negativity tolerance factor of the network
$U$ : number of prospective edges of the input
$E$ : amount of available time steps for a cycle
$F$ : number of edges to be forgotten (thus $n = F / E$ with the original notation)
$f$ : fitness factor
$a$, $b$ and $c$: relative probabilities for an edge to be positive, negative, or neutral, respectively

**Table 1.** Settings for the scale-free examination

| | Number of averaged runs | Number of points | $H$ | $U$ | $E$ | $F$ | $f$ | $a$ | $b$ | $c$ |
|---|---|---|---|---|---|---|---|---|---|---|
| Barabási model | 200 x | 10000 | – | 2 | 1 | 0 | 1 | – | – | – |

| General settings Type 1 | 200 x | 10000 | 0.5 | 2 | 10 | 1 | 1 | 1 | 1 | 1 |
|---|---|---|---|---|---|---|---|---|---|---|
| General settings Type 2 | 200 x | 10000 | 0.5 | 2 | 10 | 1 | RND | RND | RND | RND |

Obviously, if all parameters are removed we get back the Barabási model that is undoubtedly scale-free. For a general parameter set we have scale-free properties in a wide range. We do not have lower degree values with higher probabilities as there is more time ($E > 1$) to connect each input to other points. Behavior is otherwise similar to the one observed in the original Barabási model.

*Figure 2*:
The key property of a small world network is its diameter. To check whether we really have small word networks we calculated the diameter of our networks and plotted them with respect to the network size.

The original Barabási network and our model (General Settings Type 2) are represented in Fig. 2 (parameters are listed in Table 1). Please note that the scale of the plot is log–lin so the diameter is approximately a logarithmic function of network size. Our network seems to be an even smaller world than the Barabási network. The cause is simple: in this general case extra time is given for structuring that enables points to collect more links than in the Barabási model.

*Figure 3*:
As mentioned afore, if we deal with inhomogeneous inputs, then some points may obtain outstanding significance. In this simulation the fitness factor of a point is different from the others. (As earlier points usually become big centers, we performed two simulations. In the first run the special point was the first, in the second run the special point was the 32$^{nd}$. Thus, we see that in these simulations conspicuous effects occur mainly due to the changed fitness factors, and not the early integration.) The network was expanded to 1000 points.

**Table 2.** Settings for the "star shaped network" examination (the role of the fitness factor)

|  | $H$ | $U$ | $E$ | $F$ | $f$ | $a$ | $b$ | $c$ |
|---|---|---|---|---|---|---|---|---|
| Special point (1$^{st}$ or the 32$^{nd}$) | 0.5 | 1 | 10 | 1 | 3 | 1 | 1 | 1 |
| Other points |  |  |  |  | RND | RND | RND | RND |

An average of 10000 simulations is shown on Fig. 3.

As it is unambiguous from Fig. 3, an outstanding fitness factor creates a distinct position for the exceptional vertex. It gains a huge amount of links (in this extreme case more than half of the points are linked to the special point), far more than others have – this leads to its disjoint situation at the high degree end in the degree distribution. In the same time, others lose linking opportunities that causes decline at the high degree end of the distribution. Links missing here are responsible for the insufficient behavior.

*Figure 4:*

General Settings Type 2 and the original Barabási model were used to obtain Fig. 4 and prove that earlier points are of high importance. From the graph one can see that in the Barabási case the statement exactly holds (the first three points must have the same importance due to symmetrical reasons). In our model, the vast majority of points show the prescribed behavior. The first few points are of lower importance as time given to process them is not effective: all links are built and their degree can not grow further. If we reduced $E$ in the simulation, the prescribed behavior would be extended. (In this parameter set the first two points must have the same degree due to symmetrical reasons, which is correctly retained.)

*Figure 5*:
Similar effects can be reproduced to those of star shaped networks' due to high fitness factors, if extremely long processing time is given for a special point, while other parameters are unchanged.

**Table 3.** Settings for the "star shaped network" examination (time dependency)

|  | $H$ | $U$ | $E$ | $F$ | $f$ | $a$ | $b$ | $c$ |
|---|---|---|---|---|---|---|---|---|
| Special point (last) | 0.5 | 1 | 1000 | 1 | 1 | RND | RND | RND |
| Other points |  |  | 10 |  |  |  |  |  |

*Figure 6*:
Again, we used General Settings Type 2. The "attacker point" under investigation had the same parameters as the others, except for its processing time $E = 100$ and fitness factor $f = 1$.

*Figure 7*:
We used a basic network of 1000 points and in each run added a different number of new points in 1000 time steps. Fig. 7 shows the final number of points in the network. Standard deviations are marked to characterize uncertainties. We used a high $F$ parameter (forgetting) to get this curve. Settings are given in Table 4.

**Table 4.** Settings for the study on learning

|  | $H$ | $U$ | $E$ | $F$ | $f$ | $a$ | $b$ | $c$ |
|---|---|---|---|---|---|---|---|---|
| Base network (1000 points) | 0.5 | 1 | 2 | 0 | 1 | 1 | 0 | 0 |
| New information |  |  | variable | 10 |  |  |  |  |